%% file: conference_101719.tex
\documentclass[conference]{IEEEtran}
\IEEEoverridecommandlockouts
\usepackage{cite}
\usepackage{graphicx}  % add this to the preamble
\usepackage{hyperref}
\hypersetup{
    colorlinks=false,   % Disable link colors
    pdfborder={0 0 0}   % No border around links
}
\usepackage{amsmath,amssymb,amsfonts}
\usepackage{array}
\usepackage{graphicx}
\usepackage{textcomp}
\usepackage{xcolor}
\usepackage{amsmath}
\usepackage{tikz}
\usepackage{nth}
\usepackage{subcaption} 
\usepackage{fancyhdr}
\usepackage{float}%exact location 
\usepackage{multirow} % Add this package for multirow functionality
\usepackage{algorithmicx}
\usepackage{microtype}

\usepackage[linesnumbered, ruled, vlined]{algorithm2e}

\usepackage{siunitx,array,multirow}
\usepackage{booktabs}

\usepackage{tabularx}
\usepackage{multirow}
\usepackage{threeparttable}
\usepackage{multicol}
\usepackage{xcolor,colortbl}
\usepackage{amsmath}
\usepackage{makecell} % Include this package for breaking cell content into multiple lines

\usepackage[english]{babel}
\usepackage[utf8]{inputenc}
\usepackage[noend]{algpseudocode}

\newcommand{\subsubsubsection}[1]{\paragraph{#1}\mbox{}\\}
\usetikzlibrary{shapes,arrows}
\usepackage{verbatim}

\usepackage{url}
\usepackage{amsmath}
\usepackage{textcomp}
\usepackage{siunitx}
\usepackage[utf8]{inputenc}
\usepackage{upgreek}

\makeatletter
 \let\old@ps@headings\ps@headings
 \let\old@ps@IEEEtitlepagestyle\ps@IEEEtitlepagestyle
 \def\confheader#1{%
 \def\ps@IEEEtitlepagestyle{%
 \old@ps@IEEEtitlepagestyle%
 \def\@oddhead{\strut#1\hfill\strut}%
 \def\@evenhead{\strut\hfill#1\hfill\strut}%
 }%
 \ps@headings%
 }
 \makeatother   

  \confheader{
    \noindent  % Ensures no indentation
    \fontsize{9}{11}\selectfont  % Font size 9 with 11pt baseline skip
    \parbox{\textwidth}{ % Wraps text properly
        2025 International Conference on Electrical, Computer and Communication Engineering \\ 
        (ECCE), 13-15 February 2025, CUET, Chattogram-4349, Bangladesh
    }
}

  \usepackage[pscoord]{eso-pic}
 \newcommand{\placetextbox}[3]{
  \setbox0=\hbox{#3}
 \AddToShipoutPictureFG*{ \put(\LenToUnit{#1\paperwidth},\LenToUnit{#2\paperheight}){\vtop{{\null}\makebox[0pt][c]{#3}}}
  }
  }%978-1-6654-0906-3/21/$31.00
  \placetextbox{.44}{0.08}{\small{Authors' version of the paper published in proceedings of ECCE, DOI: https://doi.org/10.1109/ECCE64574.2025.11013422}}
    
\begin{document}
\title{Skin Lesion Classification Using a Soft Voting Ensemble of Convolutional Neural Networks} 
\author {\IEEEauthorblockN{Abdullah Al Shafi\IEEEauthorrefmark{1}, Abdul Muntakim\IEEEauthorrefmark{2}, Pintu Chandra Shill\IEEEauthorrefmark{3}, Rowzatul Zannat\IEEEauthorrefmark{4}, and Abdullah Al-Amin\IEEEauthorrefmark{5}}
\IEEEauthorblockA{
Department of Computer Science and Engineering, Khulna University of Engineering \& Technology, Khulna-9203, Bangladesh\IEEEauthorrefmark{1}\IEEEauthorrefmark{2}\IEEEauthorrefmark{3}\IEEEauthorrefmark{4}\\
Department of Computer Science and Engineering, Daffodil International University, Bangladesh\IEEEauthorrefmark{5}\\
abdullahratulk@gmail.com\IEEEauthorrefmark{1}, basitmuntakim@gmail.com\IEEEauthorrefmark{2}}, pintu@cse.kuet.ac.bd\IEEEauthorrefmark{3}, w.rzrowza@gmail.com\IEEEauthorrefmark{4}, alamin.cse@diu.edu.bd\IEEEauthorrefmark{5} }

\maketitle
\begin{abstract}
% Skin cancer can be identified through dermoscopic examination and ocular inspection, but early detection significantly increases survival chances. Artificial intelligence, using annotated skin images and Convolutional Neural Networks (CNNs), can improve diagnostic efficiency by enhancing the accuracy of findings. This paper presents an early skin cancer classification method that utilizes rebalanced dataset, data augmentation, filtering, segmentation, and an ensemble of CNNs. Three benchmark datasets—namely, HAM10000, ISIC 2016, and ISIC 2019—were used to train and test the proposed method. The process was initiated through rebalancing and image augmentation, followed by a series of filtering techniques. A hybrid dual encoder was developed for segmentation using transfer learning. By accurately segmenting lesion areas, the approach ensured that the subsequent classification models focus on clinically significant features, reducing the influence of background artifacts and improving classification accuracy. Classification was then performed through an ensemble of MobileNetV2, VGG19, and InceptionV3. The primary motivation for this ensemble was to strike a balance between accuracy and speed, thereby enhancing flexibility in real-world deployment. The ensemble method successfully achieved lesion recognition accuracies of 96.32\%, 90.86\%, and 93.92\% for the three datasets, respectively. The classification performance of the system was evaluated using established measures for skin lesion detection and recognition, yielding impressive results.
Skin cancer can be identified through dermoscopic examination and ocular inspection, but early detection significantly increases survival chances. Artificial intelligence (AI), leveraging annotated skin images and Convolutional Neural Networks (CNNs), enhances diagnostic accuracy. This paper presents an early skin cancer classification method using a soft voting ensemble of CNNs. Three benchmark datasets—HAM10000, ISIC 2016, and ISIC 2019—were used in this research. The process involved rebalancing, image augmentation, and filtering techniques, followed by a hybrid dual encoder for segmentation via transfer learning. Accurate segmentation focused classification models on clinically significant features, reducing background artifacts and improving accuracy. Classification was performed through an ensemble of MobileNetV2, VGG19, and InceptionV3, balancing accuracy and speed for real-world deployment. The method achieved lesion recognition accuracies of 96.32\%, 90.86\%, and 93.92\% for the three datasets. The system’s performance was evaluated using established skin lesion detection metrics, yielding impressive results.
\end{abstract}

\begin{IEEEkeywords}
Convolution Neural Networks(CNNs); Transfer Learning; Deep Neural Network; Skin Lesion Classification; Segmentation; Filtering; Ensemble Network
\end{IEEEkeywords}

\section{Introduction}
% The skin, which covers most of the body, is the largest organ of the human body and consists of up to seven layers of ectodermal tissue. Skin cancer, a common type of cancer, appears in the epidermal layer as a result of UV radiation exposure\cite{Wikipedia}. Timely diagnosis is critical for successful treatment. The increasing prevalence of skin cancer and the demand for cost-effective healthcare solutions have driven the development of computer-aided detection systems.

Comprising up to seven layers of ectodermal tissue, the skin is the biggest organ in the human body and covers the majority of the body. Skin cancer is a kind of common cancer that appears in the epidermal layer due to UV radiation exposure\cite{Wikipedia}. A timely diagnosis is essential for successful treatment. The prevalence of skin cancer and the need for low-cost health care solutions have accelerated the development of computer-aided detection systems.

Because many types of lesion images share similar visual characteristics\cite{Wikipedia}, automated skin lesion classification remains challenging. It is difficult to distinguish lesions due to variations both within and across classes. Fine-grained global context information is necessary to identify skin lesions. The use of deep learning techniques in dermatology has increased, but obtaining study approval and gathering high-quality medical images remain difficult. Accurately classifying skin diseases is an ongoing challenge.

% For automatic skin cancer detection, various traditional machine learning techniques have been employed\cite{ozkan2017skin}, such as Support Vector Machines(SVM), Decision Trees, and K-Nearest Neighbors (KNNs). Additionally, a variety of image processing methods have been applied for the same purpose, including color segmentation and texture analysis\cite{almeida2020classification}. Conventional methods of skin lesion classification face several challenges, including a lack of annotated medical image datasets, significant variability in lesion appearance, and the need for manual feature engineering. To enable quicker and more precise diagnoses, Convolutional Neural Networks (CNNs) and transfer learning\cite{hasan2022dermoexpert} are being utilized. These methods offer advantages such as automatic feature extraction and resistance to notable changes in lesion appearance. Furthermore, hybrid approaches, such as combining traditional and deep learning methods\cite{mezghani2024skin}, and multi-modal approaches\cite{adebiyi2024accurate}, are also in use. Moreover, Generative Adversarial Networks (GANs)\cite{10416885} and other techniques have been employed to improve the accuracy of skin lesion classification.
Automated skin cancer detection is closely linked to the field of medical imaging, where several classical machine learning techniques\cite{ozkan2017skin} from earlier studies deserve recognition. To aid in detection and diagnosis, additional methods such as color segmentation and texture analysis have also been utilized. However, traditional approaches still encounter several challenges, including a shortage of annotated medical image datasets, considerable variability in lesion appearance, and the necessity for manual feature engineering. To address these issues and expedite the diagnosis process, Convolutional Neural Networks (CNNs) and transfer learning\cite{hasan2022dermoexpert} are becoming increasingly popular. Additionally, new advancements are emerging that want to integrate the advantages of deep learning models and conventional statistical techniques.\cite{mezghani2024skin}. Multi-modal\cite{adebiyi2024accurate} approaches are expected to perform well in this context. Furthermore, Generative Adversarial Networks (GANs)\cite{10416885} and other techniques have been applied to improve the accuracy of the classification of skin lesions.

% \cite{almeida2020classification}
% Despite the progress made in research and technology, there is still a long way to go in improving the classification of skin lesions from images. In this study, we introduce a novel approach using a soft voting ensemble of Convolutional Neural Networks (CNNs) to tackle this challenge. Our method was tested across multiple benchmark datasets and aims to harness the strengths of three well-known models: MobileNetV2, VGG19, and InceptionV3.

Although research and technologies have come a long way, classifying skin lesions from images is still far from being perfect. Here, we present a new strategy utilizing a soft voting ensemble of CNNs through augmentation, segmentation and transfer learning to tackle this challenge. By accurately segmenting out the lesion areas from the rest of the image, it allows subsequent classification models to focus on the relevant features, unhindered by any distracting background artifacts that would otherwise degrade classification performance. We evaluated our method over several popular benchmark datasets and aimed to leverage the strengths of the three popular models: MobileNetV2, VGG19, and InceptionV3.

% Each of these models offers distinct advantages. MobileNetV2 is lightweight, fast, and efficient, making it a great choice for environments where computational resources are limited. However, it doesn’t always achieve the same level of accuracy as more complex models. VGG19, on the other hand, is deeper and excels at capturing high-level features, but it's more resource-intensive and slower. InceptionV3 is particularly good at handling complex data, thanks to its ability to extract features at multiple scales, allowing it to identify intricate patterns within images. By combining these three models in an ensemble, we aim to strike a balance between speed and accuracy, making the system more versatile and robust. This approach not only enhances the model's overall performance but also improves its ability to generalize to new data, which is essential for real-world applications.

Each of these models has its own pros and cons. MobileNetV2 is lightweight and fast\cite{sandler2018mobilenetv2}, therefore it is best suited for mobile and edge devices where computational resources are limited. But it doesn’t always get the same degree of accuracy as more complex models. VGG19 is deeper and better at abstraction of high level features but requires more resources and is slower\cite{bansal2023transfer}. InceptionV3 accommodates feature extraction at various scales, making it highly adept at identifying and recognizing fine patterns in images\cite{thwin2024skin}. By utilizing ensemble of these three models, we aim to strike a harmony between pace and accuracy, making the system more versatile and robust. It also makes the model more capable to generalize to new data, which is vital for any real-world application.  Taken together, our approach tries to achieve both the robust and timely classification of skin lesion, making it a far more practical and useful tool for real-world medical use.

% Before applying the ensemble model, we perform several filtering steps to improve the quality of the input data. This is followed by a segmentation phase, where we use a hybrid dual encoder network based on transfer learning. This hybrid approach leverages the strengths of two different encoders to capture both local and global features, improving segmentation accuracy. By accurately isolating the lesion areas from the rest of the image, the subsequent classification models can focus on the relevant features, avoiding distractions from background artifacts and ultimately boosting classification performance. In summary, our approach seeks to improve both the accuracy and robustness of skin lesion classification, offering a more practical and effective solution for real-world medical applications.

% We apply a number of filtering steps in order to enhance the quality of the input data. Then, a segmentation phase is performed by using a hybrid dual encoder network based on transfer learning. It combines the strengths of two different encoders to capture both local and global features for better segmentation.

% The paper is structured as follows: Section II reviews previous works in this area. Section III presents the proposed methodology. Section IV describes the experimental results. Section V concludes with a summary of the findings.

% The paper is organized as follows: Section II reviews related prior work in this area. Section III describes the proposed methodology. Section IV discusses the experimental results. Section V summarizes the findings and concludes.
The structure of the paper is as follows: Section II examines relevant earlier research in this field. The suggested methodology is explained in Section III. The experimental results are discussed in Section IV. The results are compiled and concluded in Section V.

\section{Literature Review}
Recent research shows that several techniques can be applied to automatically identify skin cancer using dermoscopic images. The methods vary from traditional machine learning and computer vision to deep learning with a transfer learning and GANs approaches.

Ali Ozkan et al. \cite{ozkan2017skin} demonstrated the effectiveness of machine learning methods such as SVM, KNN, Decision Trees, ANN to pre-classify the skin lesions.
% Almeida et al. \cite{almeida2020classification} developed a strategy using statistical features, GLCM, keypoints, and color channels to classify melanoma and nevus, aiding early diagnosis with optimal classifier models.

Furthermore, DermoExpert, which was evaluated on three benchmark datasets, was proposed by Kamrul Hasan et al.\cite{hasan2022dermoexpert}. It combined segmentation, augmentation, and transfer learning with CNN and outperformed traditional models by improving AUC and classification accuracy. However, the augmentation process can create artifacts that might not be present in the original clinical situations, which leads to a biased model.

% Additionally, Myat Thwin et al. \cite{thwin2024skin} achieved remarkable results with an ensemble model that combined VGG16, Inception-V3, and ResNet-50, outperforming individual models on the ISIC 2018 dataset. Hybrid CNN architectures and transfer learning have proven effective in classification of skin lesion. 

Additionally, Myat Thwin et al.\cite{thwin2024skin} proposed an ensemble of VGG16, InceptionV3, and ResNet-50 for skin lesion classification on ISIC 2018. The model achieved 92.3\% accuracy using transfer learning and augmentation. However, it did not include a segmentation step, which could have led to background noise affecting classification. The approach also had high inference time, limiting its deployment. %Anis Mezghani et al. \cite{mezghani2024skin} combined CNNs with Support Vector Machines (SVM) on the ISIC dataset, achieving improved classification accuracy.

Mezghani et al.\cite{mezghani2024skin} proposed a hybrid approach for skin lesion diagnosis. Their study used ISIC 2019 dataset and tested various ML classifiers, such as SVM, RF, and KNN, alongside CNN architectures. The hybrid model achieved improved accuracy compared to standalone ML techniques, demonstrating the benefits of feature extraction through deep networks while leveraging ML classifiers for final prediction. However, their method lacked detailed segmentation steps and faced challenges with high computational cost, making real-time application difficult.

EFAM-Net, a skin lesion classification model proposed by Zhanlin Ji et al. \cite{10695064} utilizes attention mechanisms and feature fusion, achieving accuracies of 92.30\%, 93.95\%, and 94.31\% on the ISIC 2019, HAM10000 and a private dataset, respectively. While the method improves feature extraction, its generalizability to other datasets was not assessed. 
 
Moreover, Abdulmateen Adebiyi et al.\cite{adebiyi2024accurate} employed a multimodal learning approach that used the HAM10000 dataset
with additional patient data, achieving better classification
results than models using images individually. However, with limited diversity in the dataset
in terms of demographics and geographical regions, not suitable for the global community.

% Adebiyi et al. (2024) developed a multimodal learning approach combining HAM10000 images and patient metadata, achieving 94.11% accuracy. While their model utilized metadata for enhanced performance, it lacked real-time applicability due to computational complexity.

Addressing the class imbalance issue, Qichen Su et al. \cite{10416885} proposed the Self-Transfer GAN (STGAN), a Generative Adversarial Network designed to generate synthetic skin lesion images for data augmentation, effectively balancing the dataset with a classification accuracy of 98.23\%.

The discussed survey reveals that despite advancements in skin lesion classification, there is still room for improvement. Accurate and timely classification methods are crucial, and we suggest incorporating CNN ensembles for improved classification, which makes a balance between speed and accuracy.

\section{Proposed Methodology}
% The lesion image undergoes multiple filtering techniques during preprocessing. Subsequently, segmentation is applied to identify the affected area within the filtered image. Finally, the segmented image is passed through an ensemble of CNN models for classification.

% \begin{figure}[htbp]
% \centerline{\includegraphics[scale=.36]{proposedmethod1.png}}
% \caption{Overview of the proposed pipeline for skin lesion segmentation and classification. The pipeline includes multiple stages: image preprocessing , segmentation, and classification using an ensemble of pre-trained CNN models.}
% \label{fig:Final Model}
% \end{figure}

\begin{figure}[htbp]
\centerline{\includegraphics[scale=.32]{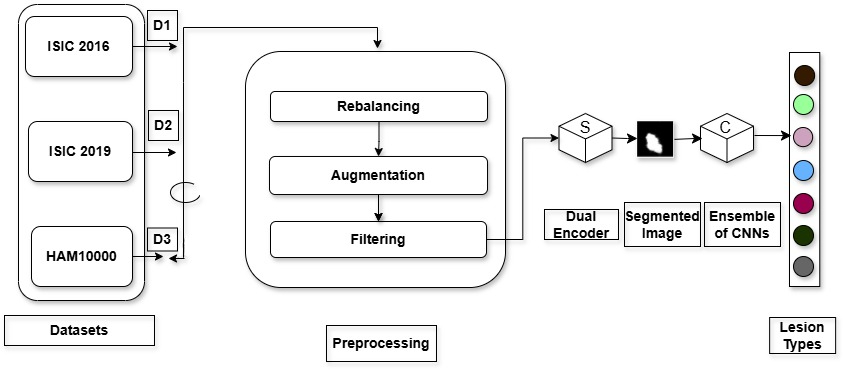}}
\caption{The complete pipeline for segmentation and classification of skin lesion. The process starts with preprocessing steps, followed by segmentation using the dual encoder model. The segmented images are then passed to an ensemble of three pre-trained models (MobileNetV2, VGG19, InceptionV3) for classification.}
\label{fig:proposed}
\end{figure}

% The basic architecture of the proposed system is shown in Fig. \ref{fig:proposed}.
The proposed system's basic design is depicted in Fig. \ref{fig:proposed}. Three distinct datasets are used: D1, D2, and D3, where D1 is a binary class dataset, and both D2 and D3 are multiclass datasets.

\subsection{Dataset}
Our proposed pipeline is trained and validated on three distinct datasets: HAM10000\cite{tschandl2018ham10000}, ISIC 2016\cite{gutman2016skin}, and ISIC 2019\cite{hernandez2024bcn20000}.

\begin{table}[ht]
    \centering
    \caption{A summary of the datasets used in the research}
    \begin{tabular}{|c|c|c|c|}
    \hline
     \textbf{Dataset} & \textbf{Lesion Types} & \textbf{Number of Images} & \textbf{Resolution}\\
    \hline
      HAM10000& 7 Classes & 10,015 & 128 × 128
    \\
    \hline
     ISIC 2016 & 2 Classes & 900 & 128 × 128
\\
    \hline
    ISIC 2019 & 8 Classes & 25,331 & 128 × 128
   \\
    \hline
\end{tabular}
\label{tab:dataset}
\end{table}

\subsection{Preprocessing}
% Image preprocessing is the initial stage in the computerized examination of images of skin lesions.
The class rebalancing, image augmentation, and filtering techniques we apply as a part of preprocessing can be succinctly summarized as follows.

\begin{itemize}
    \item Rebalancing:  The original datasets suffer from class imbalance due to variations in color tone, which is addressed through rebalancing. We applied class weights inversely proportional to class frequencies to prevent bias toward dominant classes. This ensures that misclassification penalties are higher for underrepresented classes.

\begin{figure}
     \centering
     \begin{subfigure}[b]{0.48\textwidth}
         \centering
         \includegraphics[width=\textwidth]{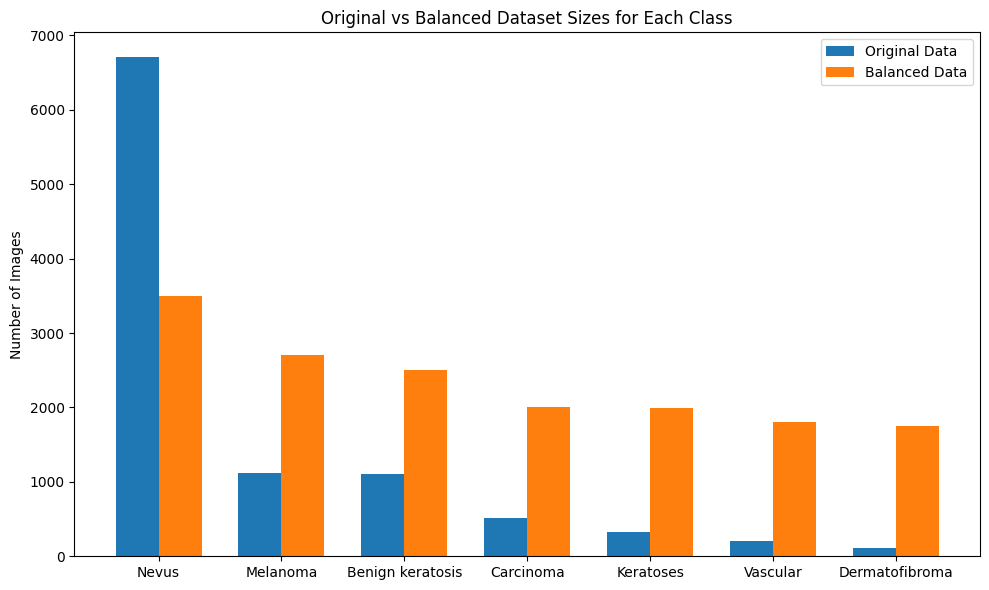}
         \caption{}
         \label{confusionisic16}
     \end{subfigure}
     \begin{subfigure}[b]{0.48\textwidth}
         \centering
         \includegraphics[width=\textwidth]{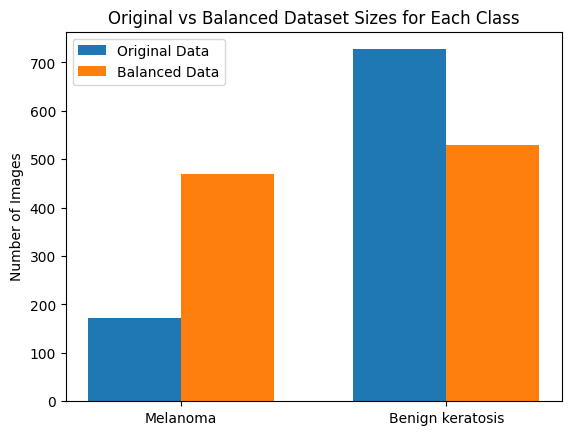}
         \caption{}
         \label{confusionisic19}
     \end{subfigure}
     \begin{subfigure}[b]{0.48\textwidth}
         \centering
         \includegraphics[width=\textwidth]{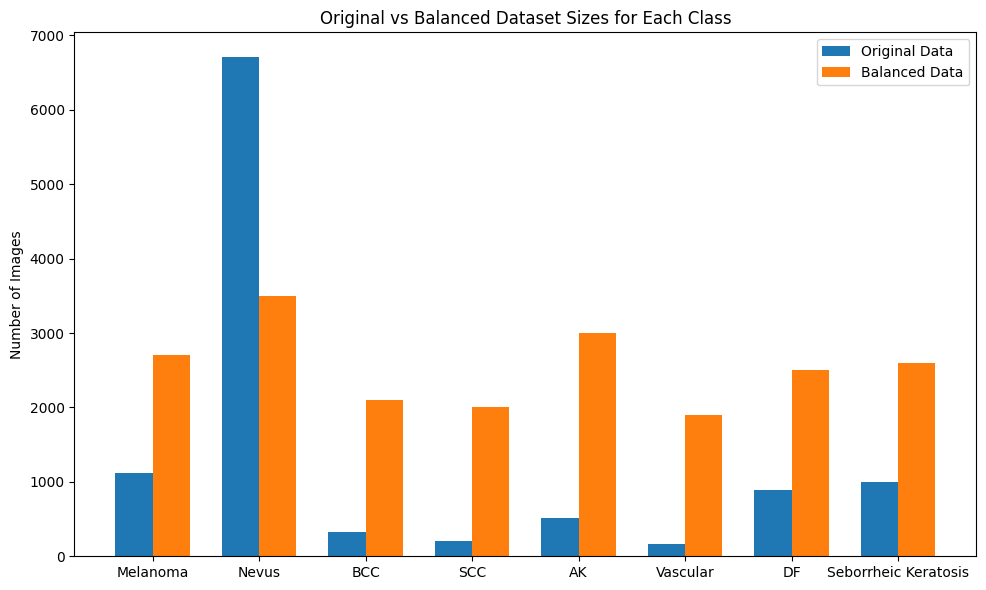}
         \caption{}
         \label{confusionham}
     \end{subfigure}
        \caption{Comparison of the original and rebalanced datasets across the (a) HAM10000, (b) ISIC-2016, and (c) ISIC-2019 datasets. The balancing process improves the class distribution, leading to more effective model training by reducing bias towards dominant classes.}
        \label{balance}
\end{figure}

\item Normalization: The next step is to normalize each image to the interval [0,1], as data normalization ensures a consistent distribution of each input parameter.
 \begin{equation}
Normalize(D) = \begin{bmatrix} 
d_{11} & \cdots  & d_{1m} \\
\vdots & \ddots & \vdots\\
a_{n1} &\cdots& d_{nm}
\end{bmatrix}/255.0
\label{elm5}
\end{equation}

\item Augmentation on minority classes: We apply affine transformations to manipulate the original image data through operations such as rotation, zooming, cropping, flipping, and translation. Figure \ref{fig:aug} shows a comparison of the original and augmented datasets.

\begin{figure}[htbp]
\centerline{\includegraphics[scale=.60]{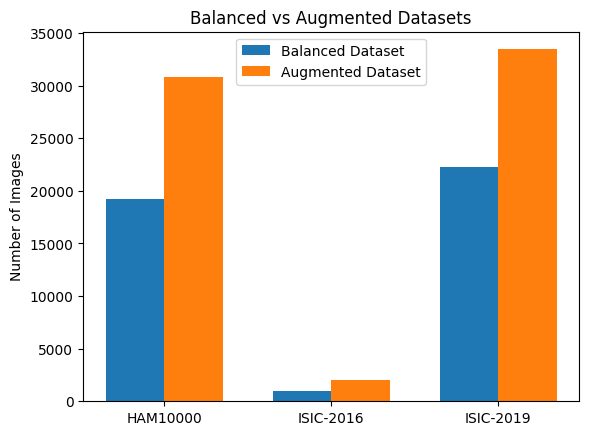}}
\caption{Comparison of Original and Balanced Datasets across HAM10000, ISIC 2016, and ISIC 2019 datasets.}
\label{fig:aug}
\end{figure}

\item Filtering: A sequence of filtering techniques, including Gaussian blur, median filter, Sobel edge detection, and histogram equalization are applied. By eliminating noise and emphasizing important aspects, this preprocessing stage improves the input image's quality and gets it ready for more precise segmentation and classification. A sample filtering result is shown in Figure \ref{fig:filtering}.

\begin{figure}[htbp]
\centerline{\includegraphics[scale=.55]{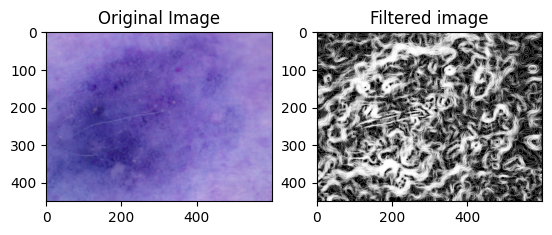}}
\caption{Filtering on a sample image, demonstrating the effect of each filter on the original image that improves clarity and contrast.}
\label{fig:filtering}
\end{figure}

\end{itemize}

\subsection{Segmentation}
\begin{figure}[htbp]
\centerline{\includegraphics[scale=.32]{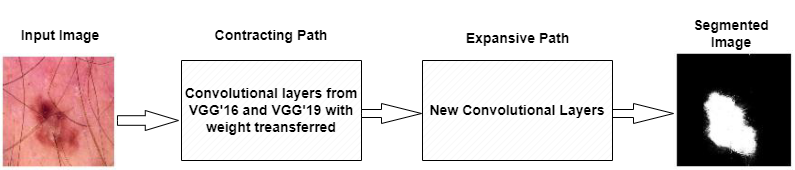}}
\caption{Illustration of the dual encoder segmentation approach. The architecture consists of two encoders, VGG-16 and VGG-19, which are used in parallel to process the input image.}
\label{fig:segmentation_process}
\end{figure}

We create a hybrid dual encoder using transfer learning. Our proposed segmentation model, shown in Figure \ref{fig:segmentation_process}, consists of two encoder instances applied to input images. The outputs of both encoders are combined to capture both fine-grained and global features, enhancing the segmentation accuracy.

\subsection{Classification Models}
\begin{algorithm}
\caption{Ensemble of m1(MobileNetV2), m2(VGG19), and m3(InceptionV3) Models}
\label{algo}
\textbf{Input:} Training data $X_{train}$, training labels $Y_{train}$, validation data $X_{val}$, validation labels $Y_{val}$, test data $X_{test}$, test labels $Y_{test}$.\\
\textbf{Output:} Ensemble accuracy (acc)
\begin{algorithmic}[1]
\Procedure{EnsembleModelTraining}{}
    \State Load the pre-trained m1, m2, and m3 models.
    \State Freeze the weights of all pre-trained models to prevent overfitting during fine-tuning.
    \State Apply global average pooling to the outputs of each model:
    \Statex \hspace{1em} $w = \text{GlobalAveragePooling}(\text{m1.output})$
    \Statex \hspace{1em} $x = \text{GlobalAveragePooling}(\text{m2.output})$
    \Statex \hspace{1em} $y = \text{GlobalAveragePooling}(\text{m3.output})$
    \State Concatenate the pooled outputs of the three models to form a single vector:
    \Statex \hspace{1em} $z = \text{Concatenate}(w, x, y)$
    \State Add a dense layer with a softmax activation function for classification.
    \State Integrate the pre-trained models into a single ensemble model:
    \Statex \hspace{1em} $ensemble_{model} = \text{CombineModels}(\text{m1.input},$ 
     \Statex \hspace{6em} $\text{m2.input}, \text{m3.input})$
     \State Compile the ensemble model Utilizing the Adam optimizer with a learning rate of 0.001 and categorical cross-entropy as the loss function.
    \State Use the training data $X_{train}$ and training labels $Y_{train}$ to train the ensemble model for 10 epochs with a batch size of 32. Validate the model with $X_{val}$ and $Y_{val}$.
    \State Make predictions on the test data $X_{test}$ using the trained ensemble model.
    \State Compute the average of the predictions:
    \Statex \hspace{2.0em} $average_{predictions} 
    = \text{AveragePredictions}(ensemble_{predictions})$
    \State Calculate the accuracy of the averaged predictions against the test labels $Y_{test}$.
    \State \textbf{return} acc \Comment{the ensemble accuracy}
\EndProcedure
\end{algorithmic}
\end{algorithm}

\subsubsection{Pretrained models}
Three different pretrained CNN architectures were used in our work.
\subsubsubsection{MobileNetV2}
\vspace*{-1mm}
% With its inverted residual structure, MobileNetV2 \cite{sandler2018mobilenetv2} achieves state-of-the-art performance in semantic segmentation and improves feature extraction and object detection by eliminating non-linearities in thin layers.
By removing non-linearities in thin layers, MobileNetV2 \cite{sandler2018mobilenetv2} enhances feature extraction and object detection while achieving cutting-edge efficiency in semantic segmentation thanks to its inverted residual structure.
% Key Features:
% \begin{itemize}
%     \item Depthwise Separable Convolutions
%     \item Linear Bottleneck
%     \item Inverted Residuals
% \end{itemize}

\subsubsubsection{VGG-19} 
\vspace*{-1mm}
% The deep CNN architecture VGG-19 \cite{simonyan2014very} consists of 19 layers, including a sequence of 3x3 convolutional filters followed by max pooling layers, and culminating in a fully connected layer.
The 19 layers that make up the deep CNN architecture VGG-19 \cite{bansal2023transfer} include a series of 3x3 convolutional filters, max pooling layers, and a fully connected layer at the end.

\subsubsubsection{InceptionV3} 
\vspace*{-1mm}
Without prior knowledge, the architecture suggests selecting the convolution type (3x3 or 5x5) for each layer. As a result, the model captures local features with fewer convolutions and high-level features with larger convolutions\cite{thwin2024skin}.

% \subsubsection{Technical Issue}
% Given InceptionV3 trained on ImageNet with 11 inception blocks or VGG19, 2 kinds of 
% experiments can be performed: 
% \begin{itemize}
%     \item Fine-tuning Inception V3 from the last 2 inception blocks.
%     \item Fine-tune the whole 
% pre-trained model. 
% \end{itemize}
% Batch Norm in Keras uses mini-batch statistics for training and inference, adjusting weights based on the new dataset's mean/variance. However, inference scales data differently due to the original ImageNet dataset, potentially affecting validation accuracy. A temporary solution is to set all layers to trainable.

\subsubsection{Proposed Ensemble Model}

In this work, we train several pre-trained models, including VGG19, MobileNetV2, and InceptionV3, on the segmented lesions. After training, a weighted average of the models' predictions is combined into an ensemble. The overall ensemble technique is illustrated in Algorithm \ref{algo}.

\section{Experimental Analysis and Results}
\subsection{Experimental Setup}

The experiments were conducted using an NVIDIA Tesla T4 GPU with 16 GB of RAM for both training and testing the models. The proposed ensemble model was implemented in Python, utilizing libraries such as Pandas, NumPy, Matplotlib, TensorFlow, and the CUDA Toolkit for GPU acceleration. The datasets were split into three subgroups: 60\% for training, 20\% for validation, and 20\% for testing. 
% The Adam optimizer, with a learning rate of 0.001 and mini-batches of 32 samples per step, was used to minimize the validation loss during training. Early stopping was applied if the validation loss did not improve after five epochs to prevent overfitting and halt training.

\subsection{Evaluation Metrics}
Accuracy, precision, recall, F1-score, and inference time are among the measures used for evaluation. Recall is the most critical metric in this context, as missing a potentially serious condition is highly risky. However, precision is also important to minimize false positives. The amount of time needed for the model to decide based on the features of the image after it has been trained is known as the inference time. Quick inference is essential for improving healthcare delivery, ensuring patient safety, and expanding access to dermatological care, particularly in remote areas.

% \begin{equation}
%      Accuracy = \frac{TP + TN}{TP + TN + FP + FN}
% \end{equation} 

% \begin{equation}
%     Precision = \frac{TP}{TP + FP}
% \end{equation}

% \begin{equation}
%     Recall = \frac{TP}{TP + FN}
% \end{equation}

% \begin{equation}
%     F_1 = 2 \cdot \frac{Precision \cdot Recall }{Precision + Recall}
% \end{equation}

\subsection{Results}
The performance of the pretrained models—MobileNetV2, VGG19, and InceptionV3—is compared to that of the proposed soft-voting ensemble model in Table \ref{result}. The ensemble model generally outperforms the individual models, particularly in terms of recall. Among the pre-trained models, InceptionV3 delivers the best overall performance, followed by VGG19, while MobileNetV2 demonstrates the lowest performance. In terms of inference time, MobileNetV2 has the shortest time, while VGG19 has the longest.

\begin{table*}[!ht]
    \centering
    \caption{The summary of the performance of the three pretrained models and the ensemble model applied to the HAM10000, ISIC 2016, and ISIC 2019 datasets. All the inference times were calculated with same experimental setup.}
    \label{result}
    %\normalsize
    \begin{tabular}{|c|c|c|c|c|c|c|}
        \hline
        \textbf{Model} & \textbf{Dataset} & \textbf{Precision} & \textbf{Recall} & \textbf{F-score} & \textbf{Accuracy} & \textbf{Inference Time}\\
        \hline
         InceptionV3 & \multirow{4}{*}{HAM10000} & 0.85 & 0.82 & 0.84 & 0.94 & 0.040 s\\
        VGG-19 &  & 0.84 & 0.81 & 0.82 & 0.93 & 0.053 s\\
        MobileNetV2 &  & 0.81 & 0.79 & 0.80 & 0.85 & 0.033 s\\
        \textbf{Ensemble} &  & \textbf{0.88} & \textbf{0.90} & \textbf{0.89} & \textbf{0.96} & 0.058 s\\
        \hline
        InceptionV3 & \multirow{4}{*}{ISIC 2016} & 0.84 & 0.83 & 0.84 & 0.89 & 0.036 s\\
        VGG-19 &  & 0.82 & 0.81 & 0.81 & 0.87 & 0.045 s\\
        MobileNetV2 &  & 0.81 & 0.79 & 0.80 & 0.83 & 0.028 s\\
        \textbf{Ensemble} &  & \textbf{0.88} & \textbf{0.84} & \textbf{0.86} & \textbf{0.91} & 0.050 s\\
        \hline
        InceptionV3 & \multirow{4}{*}{ISIC 2019} & 0.87 & 0.85 & 0.86 & 0.93 & 0.049 s\\
        VGG-19 &  & 0.84 & 0.81 & 0.83 & 0.91 & 0.060 s\\
        MobileNetV2 &  & 0.83 & 0.78 & 0.81 & 0.84 & 0.035 s\\
        \textbf{Ensemble} &  & \textbf{0.87} & \textbf{0.88} & \textbf{0.87} & \textbf{0.94} & 0.062 s\\
        \hline
    \end{tabular}
\end{table*}

\begin{figure*}[!ht]
     \centering
     \begin{subfigure}[b]{0.47\textwidth}
         \centering
         \includegraphics[width=\textwidth]{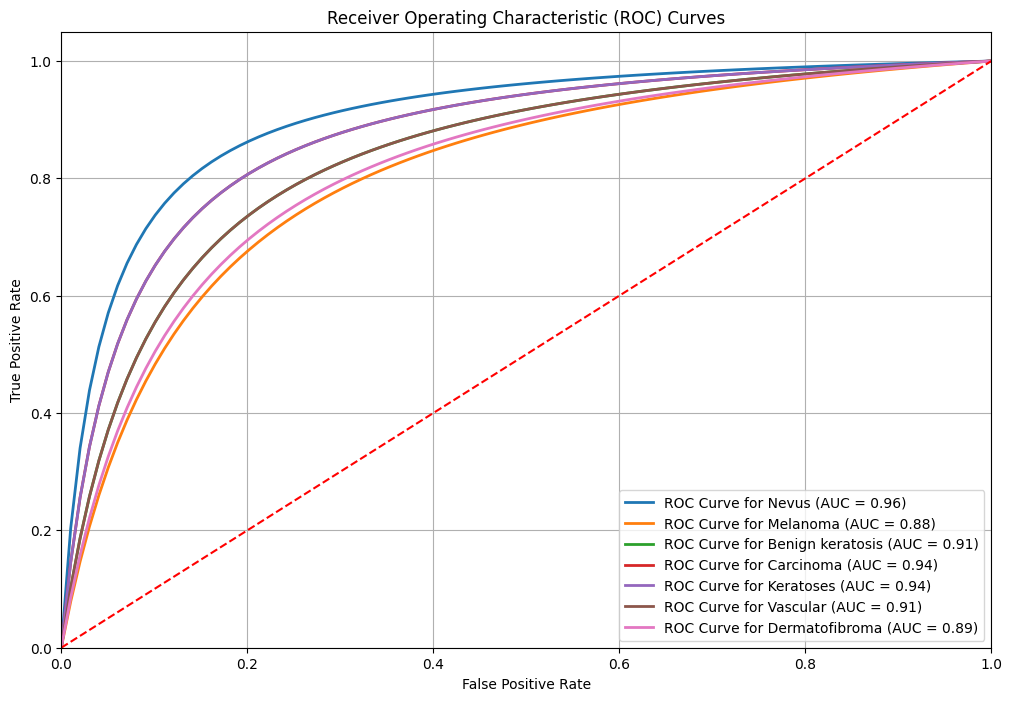}
         \caption{}
         \label{confusionisic16}
     \end{subfigure}
     \begin{subfigure}[b]{0.47\textwidth}
         \centering
         \includegraphics[width=\textwidth]{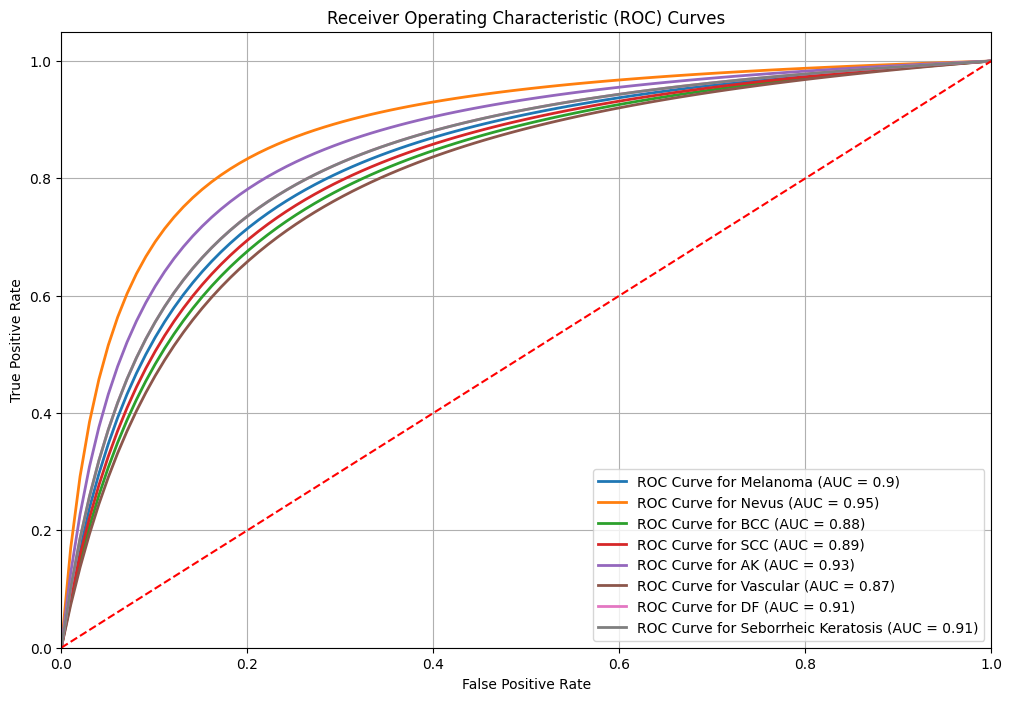}
         \caption{}
         \label{confusionham}
     \end{subfigure}
     \begin{subfigure}[b]{0.47\textwidth}
         \centering
         \includegraphics[width=\textwidth]{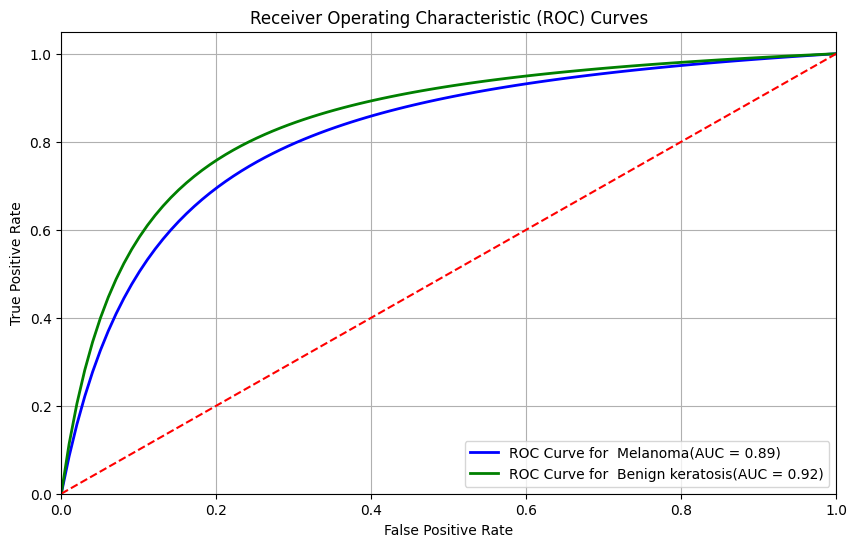}
         \caption{}
         \label{confusionisic19}
     \end{subfigure}
        \caption{ROC curves for the classification performance of the proposed method on the (a) HAM10000, (b) ISIC 2019, and (c) ISIC 2016 datasets. The curves demonstrate the model's capacity to discriminate across classes, with AUC (Area Under Curve) values ranging from 0.87 to 0.96. Lower AUC of a class suggests reduced efficiency, while the higher AUC indicates reliable differentiation from the others.
}
        \label{roc}
\end{figure*}

The significant improvement in performance from the individual pre-trained models to the ensemble model demonstrates that combining multiple models leads to a more accurate approach. The soft-voting ensemble averages predictions from three models, increasing the processing time compared to a single CNN. However, the ensemble model offers the best balance, achieving high accuracy at a reasonable inference time.
\begin{table}[htbp]
    \centering
    \caption{The comparison of the classification outcomes with some recent relevant works}
    \label{comparison}
    \begin{tabular}{|c|c|c|}
        \hline
        \textbf{Authors} & \textbf{Dataset} & \textbf{Accuracy} \\
        \hline
        Adebiyi et al.\cite{adebiyi2024accurate} & \multirow{4}{*}{HAM10000} & 94.11\% \\
        Thwin et al.\cite{thwin2024skin} &  & 90\% \\
        Ji et al.\cite{10695064} &  & 94.31\%\\
        \textbf{Proposed Model} &  & \textbf{96.32\%} \\
        \hline
        Gun et al.\cite{10757164} & \multirow{4}{*}{ISIC 2016} & 80\%  \\
        Singh et al. \cite{singh2021deep} &  & 89\% \\
        % Kaur et al. \cite{kaur2022melanoma} &  & 82\%  \\
        \textbf{Proposed Model} &  & \textbf{90.86\%} \\
        \hline
        Zhi et al. \cite{zhi2024multiclassification} & \multirow{4}{*}{ISIC 2019} & 89.04\% \\
        Ji et al. \cite{10695064} &  & 93.95\%  \\
        Gun et al. \cite{10757164} &  & 83\% \\
        \textbf{Proposed Model} &  & \textbf{93.92\%} \\
        \hline
    \end{tabular}
\end{table}

From Fig. \ref{roc}, the lower AUC values for melanoma and dermatofibroma in HAM10000 dataset suggest these classes may be less distinguishable from others. The AUC values for the two classes of ISIC 2016 are 0.89 and 0.92, supporting this observation. The lower AUCs for the vascular and BCC classes in ISIC 2019 dataset suggest reduced efficiency, while the higher AUC for Nevus indicates that this class can be reliably differentiated from the others.

Table \ref{comparison} presents a comparison between recent advanced classification approaches and our proposed solution. Our system outperforms all the methods listed in Table \ref{comparison} across all datasets. Our model demonstrates consistent performance across HAM10000, ISIC 2016, and ISIC 2019, highlighting its robustness. Furthermore, Our model maintains high recall rates, ensuring cancerous lesions are not overlooked. By incorporating MobileNetV2, we enable faster inference on mobile and edge devices, making real-time screening more accessible.

\section{Conclusions}
This study proposes a model for the classification of skin lesions utilizing a soft voting ensemble of CNNs. The method incorporates rebalanced images, data augmentation, filtering, segmentation, and an ensemble network for classification. The primary motivation behind this research was to strike a balance between accuracy and speed to enhance the system's flexibility. The evaluation of both quantitative and qualitative outcomes demonstrates how CNNs in dermatology have the potential to increase the efficacy and efficiency of diagnostic procedures. However, the system's accuracy is not yet optimal. The limitation is particularly critical if the system misclassifies a cancerous skin lesion as non-cancerous, which could lead to severe consequences. Future studies could concentrate on methods to enhance the model's classification performance, overall reliability, and efficiency in order to overcome this bottleneck.

\bibliographystyle{./IEEEtran}
% \bibliography{./IEEEexample}
\input{output4.bbl}

\end{document}

%% file: output4.bbl
% Generated by IEEEtran.bst, version: 1.12 (2007/01/11)